\def\year{2021}\relax
\documentclass[letterpaper]{article} 
\usepackage{aaai21}  
\usepackage{amssymb}  
\usepackage{amsmath}
\usepackage{times}  
\usepackage{helvet} 
\usepackage{courier}  
\usepackage[hyphens]{url}  
\usepackage{graphicx} 
\usepackage{natbib}  
\usepackage{caption} 
\usepackage{multirow}
\usepackage{booktabs}

\title{Efficient Dialogue State Tracking by Masked Hierarchical Transformer}
\author {
Min Mao,
Jiasheng Liu,
Jingyao Zhou,
Haipang Wu \\
}

\affiliations {
Hithink RoyalFlush Information Network Co.,Ltd., Zhejiang, China \\
\{maomin, liujiangsheng, zhoujingyao, wuhaipang\}@myhexin.com
}

\begin{document}

\maketitle

\begin{abstract}
This paper describes our approach to DSTC 9 Track 2:  \textit{Cross-lingual Multi-domain Dialog State Tracking}, the task goal is to build a cross-lingual dialog state tracker with a training set in rich resource language and a testing set in low resource language. We formulate a method for joint learning of slot operation classification task and state tracking task respectively. Furthermore, we design a novel mask mechanism for fusing contextual information about dialogue, the results show the proposed model achieves excellent performance on DSTC Challenge II with a joint accuracy of \textbf{62.37\%} and \textbf{23.96\%} in MultiWOZ(en $\rightarrow$ zh) dataset and CrossWOZ(zh $\rightarrow$ en) dataset, respectively.
\end{abstract}

\section{Introduction}
Task-oriented dialogue has a wide range of applications to handle everyday tasks such as booking hotels, movie tickets and restaurants, etc. The  system for supporting these tasks mostly adopts a pipelined modular architecture, which usually contains a natural language understanding (NLU) module for recognizing users' intent, a dialogue state tracking (DST) module for extracting and tracking the dialogue states, a policy (POL) module for deciding the system action and a natural language generation (NLG) module  for generating the response according to the system action. 
DST module is the core component of a task-oriented dialogue system since the system response is dependent on its result. An excellent DST can improve the user experience by reducing the number of interactions. The challenge in DSTC 9 Track 2\citep{gunasekara2020overview} is to build a cross-lingual Multi-domain DST which can track the dialogue state with resource-poor language whose original dataset is resource-rich language. The dataset of this challenge is based on MultiWOZ 2.1 \citep{eric2019multiwoz} and CrossWOZ \cite{zhu2020crosswoz}. Competitors should track the dialogue state in Chinese with MultiWOZ 2.1 dataset and in English with CrossWOZ dataset, respectively.

To solve the above tasks, we propose a multi-task model to predict the dialogue state and the state operation at each turn. The main contributions are as follows:
\begin{quote}
	\begin{itemize}
		\item To simplify the problem we use machine translation systems such as Google Translate, Baidu Translate to translate the training dataset from resource-rich language to resource-poor language, thereby the task can be considered as a traditional DST problem whose train and test dataset in the same language.   
		\item The dialogue state extracts from the dialogue history whose text usually has more than 512 tokens, but the encoder stack of Transformer usually has the length limit(512 tokens). We propose a general method to fusion the features from any length of historical turns by defining different masked self-attention structures in the transformer network. Furthermore, we use this feature fusion method to extract the global context information and local context information, respectively, merge those two representations to predict the dialogue state.     
		\item We adjust the model construction which is proposed in \citet{shan2020contextual} by adding a masked hierarchical transformer module, due to its local information module only contains the feature from a single turn.  The added module can merge multi-turn features as the local representations so that the local representations can be used to predict multi-class classification such as state operation prediction. We use a three-class state operation prediction as to the auxiliary task. However, that is a bi-class classification task in \citet{shan2020contextual} due to the local representation module in their method only contain the information from the current turn. 
		\item	Exploring data augmentation to improve the model performance.
	\end{itemize}
\end{quote}
\begin{table*}[]
	\centering
	\resizebox{\textwidth}{!}{ 
	\begin{tabular}{l}
		\hline
		\begin{tabular}[c]{@{}l@{}}\textbf{User}: I want to find a cheap restaurant in the north part of town. What is the address and food ?\\ \textbf{State}: \textit{price} range=cheap\end{tabular}                                                                                                                                                                                                            \\ \hline
		\begin{tabular}[c]{@{}l@{}}\textbf{Sys}: There is \textbf{\underline{Royal Spice which serves Indian food }}and \textbf{\underline{Da Vinci Pizzeria which serves Italian food}}. Would you like me to book you a table at either of these restaurants ?\\ \textbf{User:} \textbf{\underline{Royal Spice }}sounds good. Can you reserve a table for 8 at 15:45 on Saturday ?\\ \textbf{State: }\textit{price range=cheap; restaurant-name=Royal Spice; \underline{\textbf{food=Indian}}}\end{tabular} \\ \hline
		\begin{tabular}[c]{@{}l@{}}\textbf{Sys}: Unfortunately , i was not able to book for that specific time and day . Can i try a different day or time for you ?\\ \textbf{User:} How about at 14:45 ?\\ \textbf{State}: \textit{price range=cheap; restaurant-name=Royal Spice; food=Indian}\end{tabular}                                                                                                                 \\ \hline
		\begin{tabular}[c]{@{}l@{}}\textbf{Sys}: Sorry, the booking was unsuccessful. would you like for me to find another restaurant ?\\ \textbf{User:} Please tell me \textbf{\underline{the address of Da Vinci Pizzeria}}.\\ \textbf{State}: \textit{price range=cheap; restaurant-name=Da Vinci Pizzeria; \underline{\textbf{food=Italian}}}\end{tabular}                                                                                                       \\ \hline
	\end{tabular}}
	\caption{An example dialogue. At the last turn(the \textbf{$4$-th} turn), the underlined value of slot ``food" is corrected by the information at the \textbf{$2$-nd} turn.}
	\label{example}
\end{table*}

\section{Related Work}
Traditional DST methods can be divided into two major types: open-vocabulary\citep{le2020non,goel2019hyst, wu2019transferable} and predefined-ontology\citep{lee2019sumbt,shan2020contextual}. The former one generates slot value at each turn by a generative model such as the decoder stack in RNN and Transformer, and the latter predefines the dialogue ontology and simplifies the DST models into a classification problem. The open-vocabulary methods can partly track unseen slot values but usually has a lower performance than the predefined-ontology methods. Since the ontology in the \textit{ninth DSTC Track 2 Cross-lingual Dialog State Tracking Task} is predefined, we here use the Predefined-ontology methods aims to achieve better performance.

On the other hand, traditional DST models \citep{henderson2014word,chao2019bert} usually neglect the dialogue history and consider only utterances at current turn. To avoid the problem of lacking historical context, recent researchers employ autoregressive models to extract historical information. Some of them use a low-level network such as RNN, GRU to interactions between context and slots\citep{lee2019sumbt,goel2019hyst}, others use partial context only\citep{kim2019efficient,sharma2019improving}. These methods cannot extract the relevant context in an effective way.

Since the transformer network has been proposed in 2017\citep{vaswani2017attention}, the large-scale pre-trained model such as BERT\citep{devlin2018bert}, RoBERTa\citep{liu2019roberta} demonstrate a strong effect on NLP tasks. However,
due to the maximum sequence length limit(eg, 512 for BERT-Base), these models unable to tackle sequences that are composed of thousands of tokens. We here use the feature fusion method on dialogue history, the method can fuse any partial history information through a predefined mask in the transformer network. Furthermore, we consider the historical dialogue information as global information and utterance at the current turn and its adjacency dialogue history as the local information which is in contrast to most existing DST methods depending on either local or global information only. Whilst the local feature aims to predict state operations (UPDATE, CARRYOVER, DONTCARE ) and the global feature exploits relevant context from dialogue history. To the end, we formulate a two-branch architecture, with one branch for learning localized state operation and the other for learning slot value extraction. The two branches are not independent but synergistically jointly learned concurrently. We wish to discover and optimize jointly correlated complementary feature selections in the local and global representations. 

\section{Approach}
\subsection{Problem definition}
We assume a dialogue with $T$ turns $D=\left\{ \left( A_{1},U_{1}\right)  ,...,\left( A_{T},U_{T}\right)  \right\}$ where $A_{t}$ denotes Agent response, $U_{t}$ denotes user utterance at turn $t$, the predefine ontology as $O=\left\{ \left( s,v_{s}\right)  ,s\in S,v_{s}\subset V\right\}$ where $S$ is the set of slot names, here the slot name is denoted as domain-slot, for example, ``restaurant-name". $V$ is the total set of slot values and $v_{s}$ is the set of slot values belong to slot $s$, i.e. $v_{s}\subseteq V$, we define $B_{t}=\left\{ \left( s^{j}_{t},v^{j}_{s,t}\right)  ,1 \leqslant j \leqslant J\right\} $ as the belief state at each turn $t$, where $J$ is the total number of slots. We add ``none" to the no value slot at current turn. 

\begin{figure*}[t]
	\centering
	\includegraphics[width=0.8\textwidth]{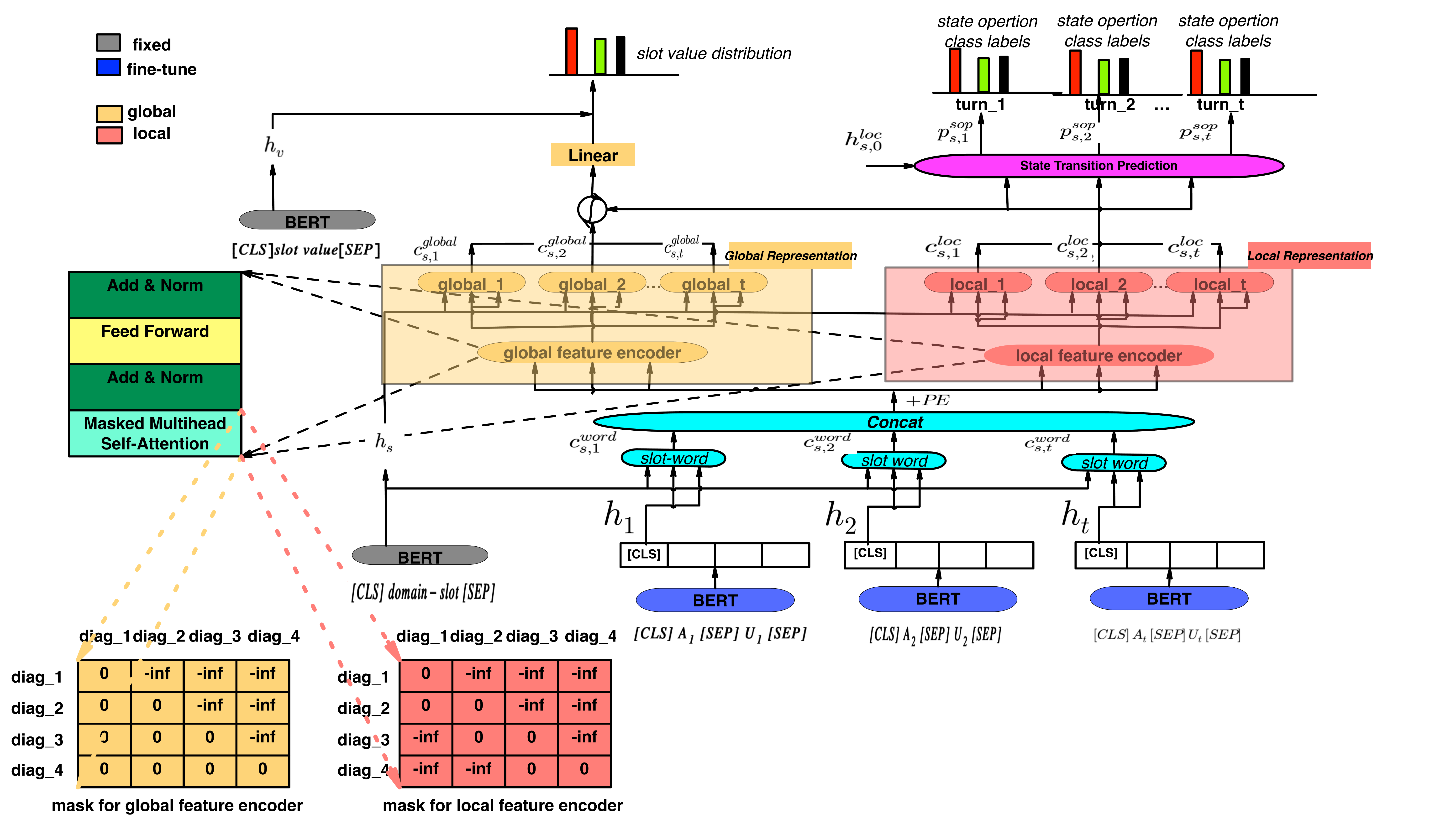} 
	\caption{The architecture of our model. We utilize the Masked  Hierarchical Transformer to encode the global and local information, respectively. We combine the global and local information to predict slot values and use the local information to predict state transition at turn $t$. Finally, we train both tasks jointly.}
	\label{fig1}
\end{figure*}

\subsection{Joint Learning Multi Loss }
Figure \ref{fig1} shows the design of the proposed model architecture. The joint learning model consists of two branches Transformer network: (1) The local branch learning the state transition of each slot; (2) Another global branch responsible for learning the slot value label of each slot at each turn. For discovering correlated complementary information between local and global feature selections, the joint learning scheme is considered with two principles as follows: 
\begin{quote}
	\begin{itemize}
         \item Shared low-level features. We construct the two types of branches on a shared lower BERT model. The intuition is that, the low-level features such as word and phrase representations which are common to all patterns in the same sentences. The local and global feature learning branches are two related learning tasks, sharing the low-level BERT model reduces the model overfitting risks. 
         \item Multi-task independent learning. For learning the maximizing of complementary discriminative features from local and global representations, the remaining layers of two branches are learned independently which aims to preserve both local saliencies in state operation prediction and global robustness in dialogue history representation.		
	\end{itemize}
\end{quote}

\subsection{Feature Fusion }
We encoded the turn-level sentence by the BERT-base model due to its length is usually less than 128. On the other hand, since the sentence length in context-level is usually more than 512, we use the masked hierarchical transformer to fuse the feature from each turn. In addition, we fuse total historical context as a global representation that contains all dialogue information up to now and $n$-history context as a local representation at the current turn. The local representation is also used to predict state operation at the current turn (eg. \textit{carryover, update, dontcare}). Due to the state operation cannot be decided by a single turn, we here use $n$-history context as the local representation ($n\geq1$). We also adjust the model construction in \citet{shan2020contextual} by adding a masked hierarchical transformer module after the BERT encoder.

\subsection{Network Construction}
As shown in Figure 1, the slot names and slot values are encoded by the same BERT with fixed weights. The dialogue is first encoded by a trainable BERT in turn-level and then fuse the turn-level features for global and local representations, respectively. Moreover, the global and the local representations are merged by a gate mechanism. The vectors from this gate mechanism are then used to compute the distance with slot value labels (similar to \citet{shan2020contextual}).  Furthermore, the local representations are used for predicting state operations as an auxiliary task. To make the text more aligned with Figure 1, we will describe each module in more detail.         

\subsubsection{Dialogue History Encode}
\begin{figure}[t]
	\centering
	\includegraphics[width=0.9\columnwidth]{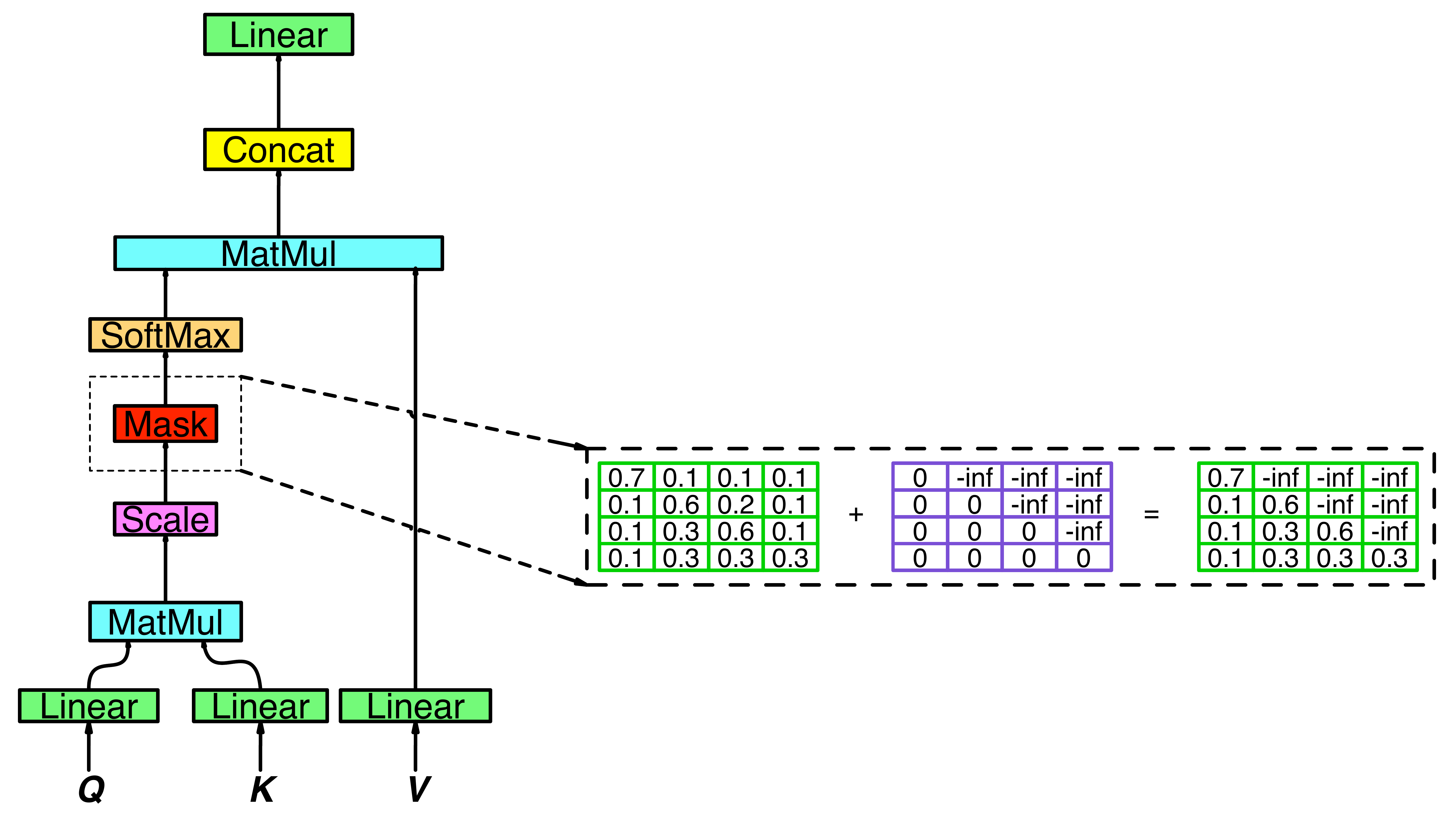} 
	\caption{Masked Transformer. The position with \textbf{-inf} value in mask matrix will be masked, since the softmax of \textbf{-inf} is \textbf{zero}.  }
	\label{fig2}
\end{figure}

As the belief state is dependent on the historical dialogue, \citet{shan2020contextual} use a masked hierarchical transformer to encode the dialogue context. We extend this method to encode the context as global and local representations with two different mask metrics. An example of Masked Self-Attention is shown in Figure \ref{fig2}.

The information of utterance at each turn is aggregated by a trainable BERT and the utterance at turn $t$ is consisted of user utterance $U_t$ and agent response $A_t$  \citep{lee2019sumbt}. We denote the turn input as $D_{t}=\left[ CLS\right]  \oplus A_{t}\oplus \left[ SEP\right]  \oplus U_{t}\oplus \left[ SEP\right] $  and the turn-level informations $h_t$ is encoded by BERT as follows:\\
\begin{equation}
h_{t}=\text{BERT}_{\text{uttr}}\left( D_{t}\right) \label{utterance_encoder}
\end{equation}

\noindent The slot name $s$ and the slot value $v$ in \citet{lee2019sumbt} are encoded by a fixed weights BERT. The sequence of slot name and slot value are denote as $q_{s}=\left[ CLS\right]  \oplus s\oplus \left[ SEP\right]$  and $q_{v}=\left[ CLS\right]  \oplus v\oplus \left[ SEP\right] $ , the outputs of token $\left[ CLS\right] $ in both $q_s$ and $q_v$ are used to represent the information of slot name and slot value, respectively.
\begin{equation}
h_{s}=\text{BERT}_{\text{slot}}\left( q_{s}\right) \label{slotname_encoder}
\end{equation}
\begin{equation}
h_{v}=\text{BERT}_{\text{slot}}\left( q_{v}\right)  \label{slotvalue_encoder}
\end{equation}

\noindent For more general situations, the slot value at a single turn cannot be discriminate only by current utterance, but is dependent on previous turns, as shown in Table \ref{example}. Furthermore,  the local feature with $n$-historical context $c^{loc}_{s,t}$ at turn $t$ can be defined as a multi-head attention between slot name and context of $n$-history, where $n$-history denotes $n$ turns of dialogue history before current turn $t$,  i.e. $\left\{ uttr_{t-n},uttr_{t-n+1},...,uttr_{t}\right\} $. Formally,  $c^{loc}_{s,t}$ can be denoted as follows:
\begin{equation}
c^{loc}_{s,t}=\text{MultiHead}\left( h_{s},c_{s,t-n\leqslant i\leqslant t},c_{s,t-n\leqslant i\leqslant t}\right)  \label{local_information_encode}
\end{equation}

\begin{figure*}[t]
	\centering
	\includegraphics[width=0.8\textwidth]{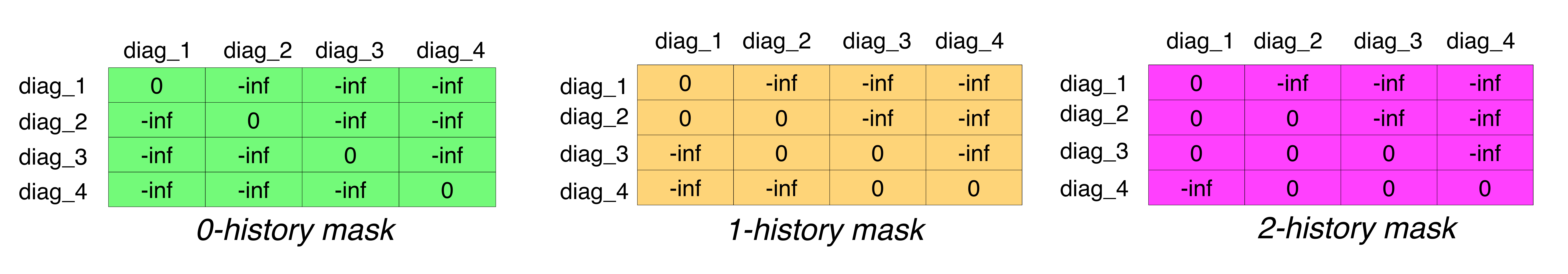} 
	\caption{$n$-history mask. The blocks with 0 score in the mask matrix means the corresponding utterance is attendable. }
	\label{fig3}
\end{figure*}

\noindent Where \textbf{$c_{s,t-n\leqslant i\leqslant t}$} is the masked hierarcharical encoder result. Figure \ref{fig3} shows what $n$-history mask matrix looks like and \textbf{$c_{s,t-n\leqslant i\leqslant t}$} is denoted as follows:

 \begin{gather}
   m^{0}=\left[ c^{word}_{s,1},c^{word}_{s,2},...,c^{word}_{s,t}\right]  \  +\  \left[ PE\left( 1\right)  ,PE\left( 2\right)  ,...,PE\left( t\right)  \right],\nonumber \\ 
    m^{N}=\ \text{MaskedTransformer}\left( m^{N-1},m^{N-1},m^{N-1}\right), \nonumber \\
   c_{s,t-n\leqslant i\leqslant t} = m^{N}      
\end{gather}  

\noindent Where $m^{N}$ is the output of $N$-layers MaskedTransformer. $PE\left( \cdot \right)$ denotes positional encoding define in \citet{devlin2018bert},  $c_{s,t}^{word}$ is the MultiHead Attention between \textit{slot name} and \textit{utterance tokens} at turn $t$, which can be defined as follows:

\begin{equation}
 c_{s,t}^{word}  = \text{MultiHead}(h_s,h_t,h_t) \label{word_information_encode}
\end{equation}

\noindent We use \textbf{$c_{s,0\leqslant i\leqslant t}$} and \textbf{$c_{s,t-n\leqslant i\leqslant t}$} are the global and local contextual information, respectively.  The feature fusion method  is shown in Figure \ref{fig1}, we expect this feature can better represent the dialogue information through balancing the information from global and local context. 

Finally, we take the method of \citet{shan2020contextual} to compute the loss for slot-value prediction, the probability distribution of slot value $p\left(v_{t}\mid U_{t},A_{t},s\right) $  at turn $t$  and its loss are defined as follows:

\begin{gather}
p\left( v_{t}\mid U_{\leq t},A_{\leq t},s\right)  \  =\  \frac{exp\left( -\| d_{s,t}-h_{v}\| \right)  }{\sum_{{v^\prime} \in V_{s}} exp\left( \| d_{s,t}-h_{v^{\prime }}\| \right)  }  \nonumber \\
L_{sv}=\sum_{s\in S} \sum^{T}_{t=1} -\text{log}\left( p\left( v_{t}\mid U_{\leq t},A_{\leq t},s\right)  \right)  
\end{gather}

\noindent Where $d_{s,t}$ denotes the encoder result with slot name $s$, it will be used to match each slot value representation $h_v$ belongs to $s$. 

\subsubsection{State Operation Decoder}
We define \textit{\textbf{O}}=\{\textit{CARRYOVER, DONTCARE, UPDATE}\}  as the state operation category set. An operation $r_t^j$ of slot $j$ is classified by state operation predictor, the mean of each category as follows:
\begin{quote}
	\begin{itemize}
		\item[] \textit{\textbf{CARRYOVER}}: slot value is unchanged.
		\item[] \textit{\textbf{UPDATE}}: slot value is changed to a new value from previous one.
		\item[] \textit{\textbf{DONTCARE}}: slot neither needs to be considered important nor be tracked in the dialogue.
	\end{itemize}
\end{quote}
		
\subsubsection{Input Representation for State Operation}

As the conversation progresses, the state operation at each turn is determined by the previous dialogue state and the current dialogue turn. The flow of state can be modeled by autoregressive decoders. Therefore, we use RNN as our decoder model,  $c_{s,t}^{loc}$ and $h_{s,t-1}^{loc}$ as the model inputs:
\begin{equation}
h^{loc}_{s,t}  = \text{RNN}(c_{s,t}^{loc},h_{s,t-1}^{loc} )
\end{equation}

\noindent Where $h_{s,t}^{loc}$ is the representation of state operation at turn $t$.

The probability distribution over state operations $p_{s,t}^{sop}$ and its loss are defined as follows:
\begin{gather}
p^{sop}_{s,t}=\text{softmax}\left( W_{sop}h^{loc}_{s,t}\right), \nonumber \\
L_{sop}\  =\  \sum_{s\in S} \sum^{T}_{t=1} -\left( Y_{s,t}^{sop}\right)^{T}  \text{log}\left( p^{sop}_{s,t}\right)  
\end{gather}
\noindent Where $W_{sop}$ is a linear project to obtain operation probability distribution $p^{sop}_{s,t}$ and  $Y^{sop}_{s,t}$ is the operation label of slot $s$ at turn $t$ .

Therefore, we take the sum of losses metioned above as the final joint loss $L_{joint}$ as following:
    \begin{equation}
    L_{joint}=\  L_{sv}+L_{sop}
    \end{equation}

\section{Experiment}
\subsection{Baseline Models}
We compare the performance of the proposed method with the following models:
\begin{quote}
	\begin{itemize}
		\item[] \textbf{SUMBT}: Using a trainable BERT to encode system and user utterances and a fixed weighted BERT to encode slot-type and slot-value information and predict the slot-value label based on a certain metric\citep{lee2019sumbt}. 
		\item[] \textbf{CHAN}: Employing a contextual hierarchical network to fusion contextual information and exploiting the same method in SUMBT to predict the slot-value label\citep{shan2020contextual}.   
		\item[] \textbf{NADST}: Generate dialogue state at each turn by a non-autoregressive decoder model\citep{le2020non}. 
		\item[] \textbf{TRADE}: Using an Encoding Decoding model to generate slot-value label\citep{wu2019transferable}. 
	\end{itemize}
\end{quote}
		
\begin{table}[]
	\resizebox{\columnwidth}{!}{ 
		\begin{tabular}{l|c|c|c}
			\toprule[2pt]
			\textbf{	\multirow{2}{*}{Model}}   & \textbf{\multirow{2}{*}{Ontology}} & \textbf{MultiWOZ(en $\rightarrow$ zh) }& \textbf{CrossWOZ(zh $\rightarrow$ en) }\\ \cline{3-4} 
			&                           & \textbf{Joint(\%) }                               & \textbf{Joint(\%)  }                              \\ \hline
			TRADE                    & $\times$                     & 29.63                                    & 7.9                                      \\ 
			NADST                    & $\times$                     & 31.21                                    & 8.3                                      \\ 
			SUMBT                    & $\checkmark$                 & 49.4                                     & 10.6                                     \\ 
			CHAN                     & $\checkmark$                 & 49.19                                    & 11.3                                     \\ \hline
			\textbf{CHAN + 4-class STP(Ours)} & \textbf{$\checkmark$}                 & \textbf{50.16 }                                   & \textbf{11.87}                                    \\
			\bottomrule[2pt]
	\end{tabular}}
	\caption{ Joint accuracy on human val sets of MultiWOZ(en $\rightarrow$ zh) and CrossWOZ(zh $\rightarrow$ en), respectively. The ontology column indicates if the model is based on predefined ontology or not.}
	\label{baseline_model_result}
\end{table}

\noindent Table \ref{baseline_model_result} shows the joint accuracy of baseline models and our model on MultiWOZ (en $\rightarrow$ zh)  and  CrossWOZ (zh $\rightarrow$ en)  human val datasets. Our model achieves 50.16\% and 11.87\% with 0.96\% and 0.57\% improvement respectively. 

\subsection{Data Augmentation}
We found that the human evaluation dataset is generated through real-life conversations, while the training data is generated by Google translator, so it is not as natural as human language. 
In this work, we consider data augmentation to improve  model performance. We use translation services provided by Tencent, Baidu, and our translation model to translate the source language to the target language. We also found that MultiWOZ(en $\rightarrow$ zh) and CrossWOZ(zh $\rightarrow$ en) are provided by different organizers, both of them have some annotation errors, so we here use different methods to correct slot value labels.

For the MultiWOZ(en $\rightarrow$ zh) dataset we using the label in MultiWOZ\_2.2 as right labels to correct the older ones, and the changes of each slot on MultiWOZ(en $\rightarrow$ zh) are shown in Table \ref{slotvalue_modified}
\begin{table}[]
	\centering
	\resizebox{0.9\columnwidth}{!}{ 
		\begin{tabular}{cccc}
			\toprule[2pt]
			\textbf{\textit{Slot name}}            & \textbf{\textit{\# of total slot values}} & \textbf{\textit{\# modified slot values}} & \textbf{\textit{\% of slot values modified}} \\ \hline
			attraction-type      &10525                         & 845                     &8                  \\ \hline
			restaurant-food      &16095                         & 502                     &3.1                   \\ \hline
			attraction-name      &5843                         & 465                     &7.9                   \\ \hline
			restaurant-name      &7293                         & 336                     &4.6                   \\ \hline
			train-leave at       &7563                         & 331                     &4.38                   \\ \hline
			hotel-name           &8621                         & 296                     &3.43                   \\ \hline
			taxi-departure       &4037                         & 254                     &6.29                   \\ \hline
			taxi-destination     &4108                         & 204                     &4.97                   \\ \hline
			train-arrive by      &7488                         & 167                     &2.23                   \\ \hline
			restaurant-book-time &8958                         & 156                     &1.74                   \\ 
			\bottomrule[2pt]
	\end{tabular}}
	\caption{The \textbf{Top-10} number of slot values modified }
	\label{slotvalue_modified}
\end{table}

For the CrossWOZ(zh $\rightarrow$ en) dataset we found that the belief states at some turns are not inherited its previous turns. We consider these as the labeling errors that need to be corrected, so we here use the concept of state transition (carryover, update, dontcare, delete) to correct the belief state at each turn. 

We estimate the effectiveness of the back-translation data augmentation and state transition prediction task.  The joint accuracy reduces by $0.58\%$ with removing the state operation prediction task and reduces by $8.4\%$ with no data augmentation. Moreover, the performance  decreases by $9.4\%$ with removing both of them.  Table \ref{ablation} demonstrates that the data augmentation and state transition prediction task are crucial for DST.

\begin{table}[]
	\resizebox{0.9\columnwidth}{!}{ 
		\begin{tabular}{lll}
			
			\toprule[2pt]
			& \textbf{MultiWOZ(en $\rightarrow$ zh)} & \textbf{CrossWOZ(zh $\rightarrow$ en)} \\ \hline		
			Final Model                        & 58.56                                    & 16.81                                    \\ \hline
			remove state transition prediction & 57.98(-0.58)                                    & 16.29(-0.52)                                    \\ \hline
			remove data argumentation          & 50.16(-8.4)                                    & 11.87(-4.94)                                    \\ \hline
			remove above two(only CHAN)                   & 49.19(-9.37)                                    & 11.3(-5.51)     \\
			\bottomrule[2pt]
	\end{tabular}}
	\caption{The ablation study of the state transition prediction and the data Augmentation on MultiWOZ(en $\rightarrow$ zh) and CrossWOZ(zh $\rightarrow$ en), respectively. }
	\label{ablation}
\end{table}

\begin{table}[]
	\centering
	\resizebox{\columnwidth}{!}{ 
		\begin{tabular}{cccccc}
			\toprule[2pt]
			\textbf{Team}     & \textbf{Joint(\%)}   & \textbf{Slot(\%) }   & \textbf{Slot P/R/F1}       & \textbf{Joint(pub/pri)} & \textbf{Rank} \\ \hline 
			\textbf{1(ours)}  & \textbf{62.37} & \textbf{98.09 }& \textbf{92.15/94.02/93.07} & \textbf{62.70/62.03}  & \textbf{1}    \\ 
			2        & 62.08 & 98.10 & 90.61/96.20/93.32 & 63.25/60.91  & 2    \\
			3        & 30.13 & 94.40 & 87.07/74.67/80.40 & 30.53/29.72  & 3    \\
			Baseline & 55.56 & 97.68 & 92.02/91.10/91.56 & 55.81/55.31  & N/A \\    
			\bottomrule[2pt]
		\end{tabular}
	}
	\caption{MultiWOZ Leaderboard (Best Submissions).}
	\label{multiwoz_leaderboard}
\end{table}

\begin{table}[]
	\centering
	\resizebox{\columnwidth}{!}{ 
		\begin{tabular}{cccccc}
			\toprule[2pt]
			\textbf{Team}     & \textbf{Joint(\%)}   & \textbf{Slot(\%)}    & \textbf{Slot P/R/F1}       & \textbf{Joint(pub/pri)} & \textbf{Rank} \\ \hline
			3        & 16.86 & 89.11 & 68.26/62.85/65.45 & 16.82/16.89  & 1    \\
			\textbf{ 1(ours) }       & \textbf{15.28} & \textbf{90.37 }& \textbf{65.94/78.87/71.82} & \textbf{15.19/15.37 } & \textbf{2}    \\
			2        & 13.99 & 91.92 & 72.63/78.90/75.64 & 14.41/13.58  & 3    \\
			Baseline & 7.21  & 85.13 & 55.27/46.15/50.30 & 7.41/7.00    & N/A \\
			\bottomrule[2pt]
		\end{tabular}
	}
	\caption{CrossWOZ Leaderboard (Best Submissions).}
	\label{crosswoz_leaderboard_ori}
\end{table}

\begin{table}[]
	\centering
	\resizebox{\columnwidth}{!}{ 
		\begin{tabular}{cccccc}
			\toprule[2pt]
			\textbf{Team}     & \textbf{Joint(\%) }  & \textbf{Slot(\%)}    & \textbf{Slot P/R/F1}       & \textbf{Joint(pub/pri)} & \textbf{Rank} \\ \hline
			2        & 32.30 & 94.35 & 81.39/82.25/81.82 & 32.70/31.89  & 1    \\
			\textbf{1(ours)}        & \textbf{23.96} & \textbf{92.94} & \textbf{74.96/83.41/78.96} & \textbf{23.45/24.47}  & \textbf{2}    \\
			3        & 15.31 & 89.70 & 74.78/64.06/69.01 & 14.25/16.37  & 3    \\
			Baseline & 13.02 & 87.97 & 67.18/52.18/58.74 & 13.30/12.74  & N/A  \\		
			\bottomrule[2pt]
		\end{tabular}
	}
	\caption{CrossWOZ Leaderboard (Updated Evaluation, Best Submissions).}
	\label{crosswoz_leaderboard_update}
\end{table}
\subsection{Overall Results}
We train the model with different $n$-history as the local information and we finally choice $1$-history as the best length for the joint learning. 

By using above improvements, we achieve the result with joint accuracy of $\textbf{62.37\%}$ and $\textbf{23.96\%}$ on MultiWOZ(en $\rightarrow$ zh) and CrossWOZ(zh $\rightarrow$ en) datasets, respectively.

With this end-to-end model, we achieve \textbf{Top 1} in MultiWOZ(en $\rightarrow$ zh) dataset, and \textbf{Top 2} in CrossWOZ(zh $\rightarrow$ en) dataset. The results of MultiWOZ(en $\rightarrow$ zh) and CrossWOZ(zh $\rightarrow$ en) tasks are shown in Table \ref{multiwoz_leaderboard} and \ref{crosswoz_leaderboard_ori}  respectively. The Organizers found that the CrossWOZ(zh $\rightarrow$ en) test data miss ``name" labels when the user accepts the attraction/hotel/restaurant recommended by the system\citep{gunasekara2020overview}. Table \ref{crosswoz_leaderboard_update} shows the updated leaderboard for CrossWOZ. Moreover, although the evaluation was updated in CrossWOZ(zh $\rightarrow$ en), our algorithm is still ranked second in the updated CrossWOZ Leaderboard, which shows that our method has better generalization ability.  

\section{Conclusion}
In this paper, we introduce a general feature fusion method  as the solution in DSTC 9 Track 2 competition, which can merge any parts of context feature from dialogue history. We also construct a multi-task network to improve the feature representation ability. Our proposed model is ranked first in MultiWOZ (en $\rightarrow$ zh) and second in CrossWOZ (zh $\rightarrow$ en) respectively. The proposed model is based on predefined ontology, and we will investigate an open-vocabulary model in the future.

\nocite{*}
\bibliography{paper_citition}

\end{document}